\documentclass[runningheads]{llncs}
\usepackage{graphicx}
\usepackage{amsmath,amssymb} 

\usepackage{epsfig}
\usepackage{algorithm}
\usepackage{algpseudocode}
\usepackage{multirow}
\usepackage{epstopdf}
\usepackage{wrapfig}
\usepackage{verbatim}
\usepackage[lofdepth,lotdepth]{subfig}
\usepackage{multicol}
\usepackage[normalem]{ulem}
\usepackage{color}
\usepackage{lipsum}
\usepackage{multicol}
\usepackage{subfig}
\usepackage{epstopdf}
\usepackage[width=122mm,left=12mm,paperwidth=146mm,height=193mm,top=12mm,paperheight=217mm]{geometry}
\usepackage{wrapfig}

\DeclareMathOperator*{\argmin}{arg\,min}

\newcommand{\ie}{\textit{i.e.}~}
\newcommand{\eg}{\textit{e.g.}~}
\newcommand{\etal}{\textit{et al.}~}

\begin{document}
\pagestyle{headings}
\mainmatter
\def\ECCV18SubNumber{1647}  

\title{Supervised Convolutional Sparse Coding} 

\titlerunning{Supervised Convolutional Sparse Coding}

\authorrunning{L. Affara, B. Ghanem, P. Wonka}

\author{Lama Affara, Bernard Ghanem, Peter Wonka}
\institute{King Abdullah University of Science and Technology (KAUST), Saudi Arabia\\
        \email{lama.affara@kaust.edu.sa, bernard.ghanem@kaust.edu.sa, 
        pwonka@gmail.com}
}

\maketitle

\begin{abstract}
   Convolutional Sparse Coding (CSC) is a well established image representation model especially suited for image restoration tasks. In this work, we extend the applicability of this model by proposing a supervised approach to convolutional sparse coding, which aims at learning discriminative dictionaries instead of purely reconstructive ones. We incorporate a supervised regularization term into the traditional unsupervised CSC objective to encourage the final dictionary elements to be  discriminative.
Experimental results show that using supervised convolutional learning results in two key advantages. First, we learn more semantically relevant filters in the dictionary and second, we achieve improved image reconstruction on unseen data.

\end{abstract}

\section{Introduction}
\label{sec:intro}

\begin{wrapfigure}{r}{0.55\textwidth}
  \begin{center}
  \vspace{-1.1cm}
    \includegraphics[width=0.55\textwidth]{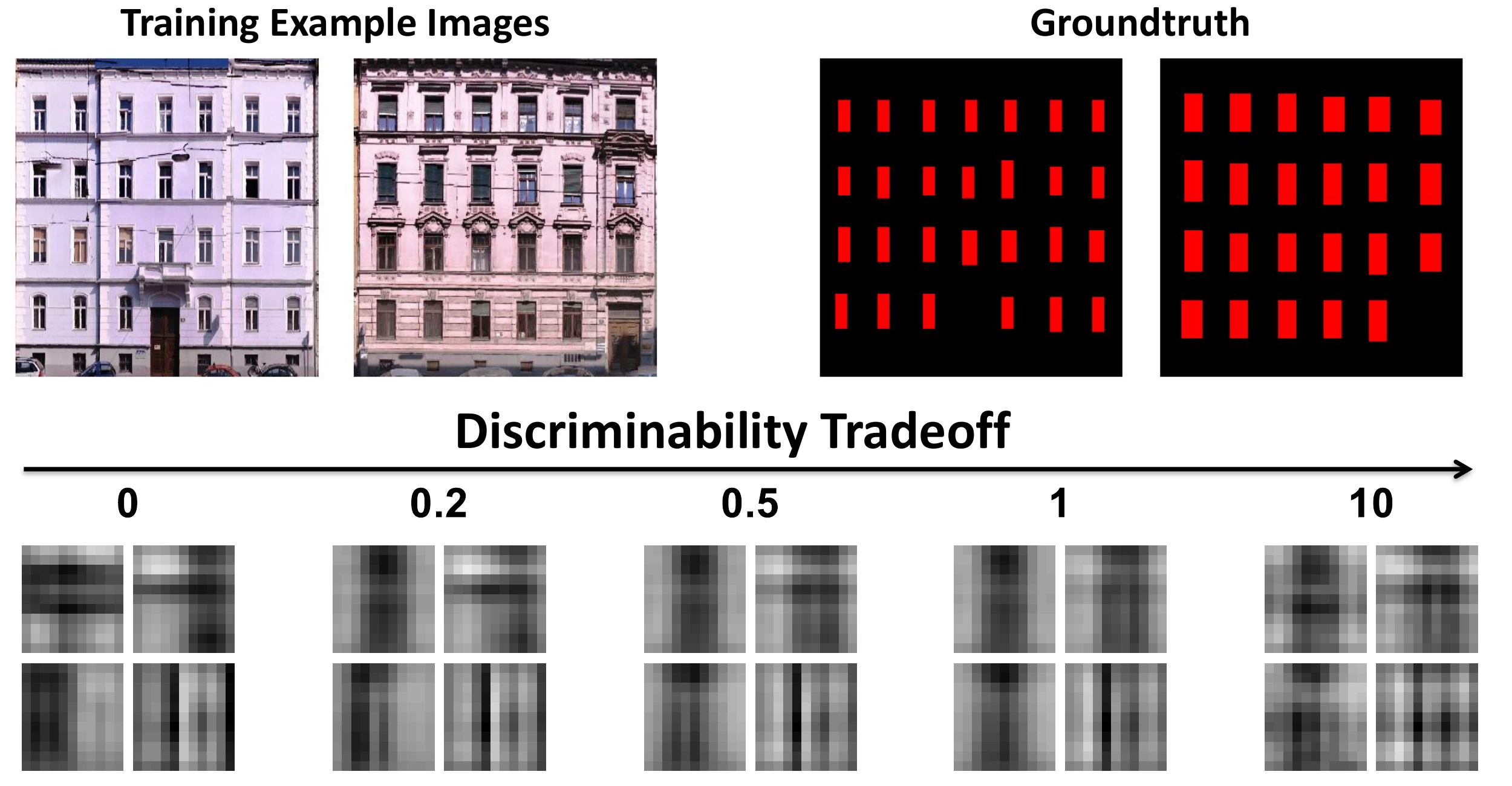}

  \end{center}
  	\caption{Top: Examples of the training and ground truth images. Bottom: Learned supervised dictionary elements as the classification regularizer is enforced more during CSC training. As the importance of the classification increases, the dictionary elements become more discriminative thus increasing their resemblance to the ground truth objects (windows in this case). }
	\vspace{-0.5cm}
	\label{fig:main}
\end{wrapfigure}
Convolutional Sparse Coding (CSC) is a rich image representation model that is inspired by the response of neurons to stimuli within their receptive fields for human vision. It is applied to various computer vision tasks such as image and video processing~\cite{elad2006image,aharon2006uniqueness,couzinie2011dictionary,yang2008image,Gu2016}, computational imaging~\cite{heide2014imaging}, line drawings~\cite{shaheen2017constrained}, tracking ~\cite{Zhang2016} as well as the design of deep learning architectures~\cite{krizhevsky2012imagenet}.

CSC is a special type of sparse dictionary learning (DL) algorithm, in which it uses the convolution operator, unlike traditional DL that uses regular linear combinations. This results in diverse translation-invariant patches while maintaining the latent structure of the underlying signal. In its formulation, CSC inherently provides sparse maps that correspond to each dictionary element. These sparse maps fire at locations where the dictionary element is prevalent.

Since  traditional dictionary learning produces dictionary elements that focus on image reconstruction (a generative task), the resulting elements might not necessarily have significant semantic value, \ie they might not be discriminative of any semantic class (\eg objects). In this context, there has been successful work \cite{mairal2009supervised,mairal2008discriminative,jiang2011learning} that studies  traditional dictionary learning from a supervised point of view.
In this case, more semantically meaningful dictionaries are generated instead of purely reconstructive ones.

Inspired by these supervised dictionary learning  approaches, we model the CSC problem as a supervised convolutional learning and coding task. We will show in this paper, that our framework leads to two important advantages. First, we can learn dictionary elements that are more semantically meaningful. Figure~\ref{fig:main} shows an example of how the semantics of window dictionary elements improves as the weight of the regularizer in our Supervised CSC framework is increased.
While semantics of patches is partially judged by visual inspection, we propose to additionally evaluate  patch semantics by comparing the classification performance of
a simple classifier trained using our supervised dictionary elements as opposed to a traditional unsupervised dictionary. This shows that our dictionary is more discriminative and therefore carries more semantic information. Second, we can improve reconstruction quality on unseen data, because the learned dictionary elements are semantically meaningful and thus  less prone to over-fitting. The surprising result of our work is that by adding a regularizer, we not only achieve a trade-off between two objectives in training, but we in fact improve both objectives simultaneously on unseen data.




Some reformulations and extensions to traditional CSC have emerged recently ~\cite{bibi2017high,choudhury2017consensus,Wang2017OnlineCS,wohlberg2016boundary}. However, there exists no prior work that handles the CSC problem from a supervised approach in which the convolutional dictionary learning incorporates groundtruth annotations of a target class. In this work, we are the first to jointly learn convolutional dictionaries that are both reconstructive and discriminative by adding a logistic discrimination loss to the CSC objective. We solve the resulting optimization in the Fourier domain using coordinate descent, since the overall objective is not jointly convex, but it is convex for each of the variables separately. We compare our approach to a baseline approach in which traditional CSC is cascaded with a classifier trained on its learned dictionary.

The rest of the paper is divided as follows. In Section~\ref{sec:related}, we discuss related work to CSC and Supervised DL. We then introduce traditional CSC in Section~\ref{sec:csc}. Section~\ref{sec:scsc} shows our formulation and optimization approach to supervised CSC. Finally,  results are presented in Section~\ref{sec:results}.

\section{Related Work}
\label{sec:related}
In this section, we show research related to the CSC problem and supervised dictionary learning since our work is most related to these two  areas in computer vision.\\
\noindent \textbf{Convolutional Sparse Coding.}~CSC has many applications and quite a few methods have been proposed for solving the optimization problems in CSC.
The seminal work of~\cite{Zeiler2010} proposes \emph{Deconvolutional Networks}, a learning framework based on convolutional decomposition of images under a sparsity constraint. Unlike previous work in sparse image decomposition~\cite{olshausen1997sparse,lee2006efficient,mairal2009online,mairal2009supervised} that builds hierarchical representations of an image on a patch level, Deconvolutional Networks perform a sparse decomposition over whole images. This strategy significantly reduces the redundancy among filters compared with those obtained by the patch-based approaches. Kavukcuoglu \etal~\cite{kavukcuoglu2010learning}
propose a convolutional extension to the coordinate descent sparse coding algorithm~\cite{li2009coordinate} to represent images using convolutional dictionaries for object recognition tasks. Following this path, Yang \etal~\cite{yang2010supervised} propose a supervised dictionary learning approach to improve the efficiency of sparse coding.
To efficiently solve the optimization problems in CSC, most existing approaches transform the problem into the frequency domain. Bristow \etal~\cite{Bristow2013} propose a quad-decomposition of the original objective into convex subproblems and they exploit the Alternating Direction Method of Multipliers (ADMM) approach to solve the convolution subproblems in the Fourier domain. In their follow-up work~\cite{Bristow2014}, a number of optimization methods for solving convolution problems and their applications are discussed. In the work of~\cite{Kong2014}, the authors further exploit the separability of convolution across bands in the frequency domain. Their gain in efficiency is due to computing a partial vector (instead of a full vector). To further improve efficiency, Heide \etal~\cite{Heide2015} transform the original constrained problem into an unconstrained problem by encoding the constraints in the objective. The new objective function is then further split into two subproblems that are easier to optimize separately. They also devise a more flexible solution by adding a diagonal matrix to the objective function to handle the boundary artifacts. Recent work \cite{Wang2017OnlineCS,wohlberg2016boundary} has also reformulated the CSC problem by extending its applicability to higher dimensions ~\cite{bibi2017high} and to large scale data~\cite{choudhury2017consensus}. We adopt the optimization strategy used in~\cite{Heide2015} as it is the state of the art unsupervised CSC solver.\\
\noindent \textbf{Supervised Dictionary Learning.}~Supervised dictionary learning (SDL) has received great attention in both the computer vision and image processing communities. SDL methods exploit the sparse signal decompositions to learn representative dictionaries whose elements can be used to discriminate certain semantic classes from each other. The algorithms vary in the procedure of supervision and the elements involved in the learning task. In the most simplistic approach to SDL, multiple dictionaries are computed for each class \cite{wright2009robust} and then combined into one \cite{varma2005statistical}. A more robust approach involves a joint learning of dictionary elements and classifier parameters to learn discriminative dictionaries\cite{mairal2009supervised,mairal2008discriminative,zhang2010discriminative}. Zhang \etal~\cite{zhang2013simultaneous} follow a similar approach where they learn discriminative projections along with the dictionary thus learning the dictionary in a projected space. More recent work \cite{bahrampour2016multimodal} formulate SDL by a multimodal task-driven approach using sparsity models under a joint sparsity prior. Yankelevsky \etal~\cite{yankelevsky2017structure} use two graph-based regularizations to encourage the dictionary atoms to preserve their feature similarities.
In \cite{wang2017cross}, the authors structure the learned dictionaries by crosslabel suppression with group regularization, which increases the computational efficiency without sacrificing classification. In this work, we are the first to embed such supervision into the CSC model, leading to convolutional dictionary elements that are more semantically relevant to the classes they are trained to discriminate.


\vspace{-0.2cm}
\section{Convolutional Sparse Coding}
\vspace{-0.2cm}
\label{sec:csc}
In this section, we present the mathematical formulation of the CSC problem and discuss state-of-the-art optimization methods to compute an approximate solution. There are multiple slightly different, but similar formulations for the CSC problem. We follow the formulation provided by Heide \etal~\cite{Heide2015} in which boundary handling is augmented in the reconstructive term and a stationary solution is guaranteed by a coordinate descent approach.


\subsection{Unsupervised CSC Model}
\begin{figure*}[t]
	\centering
		\includegraphics[width=1\linewidth]{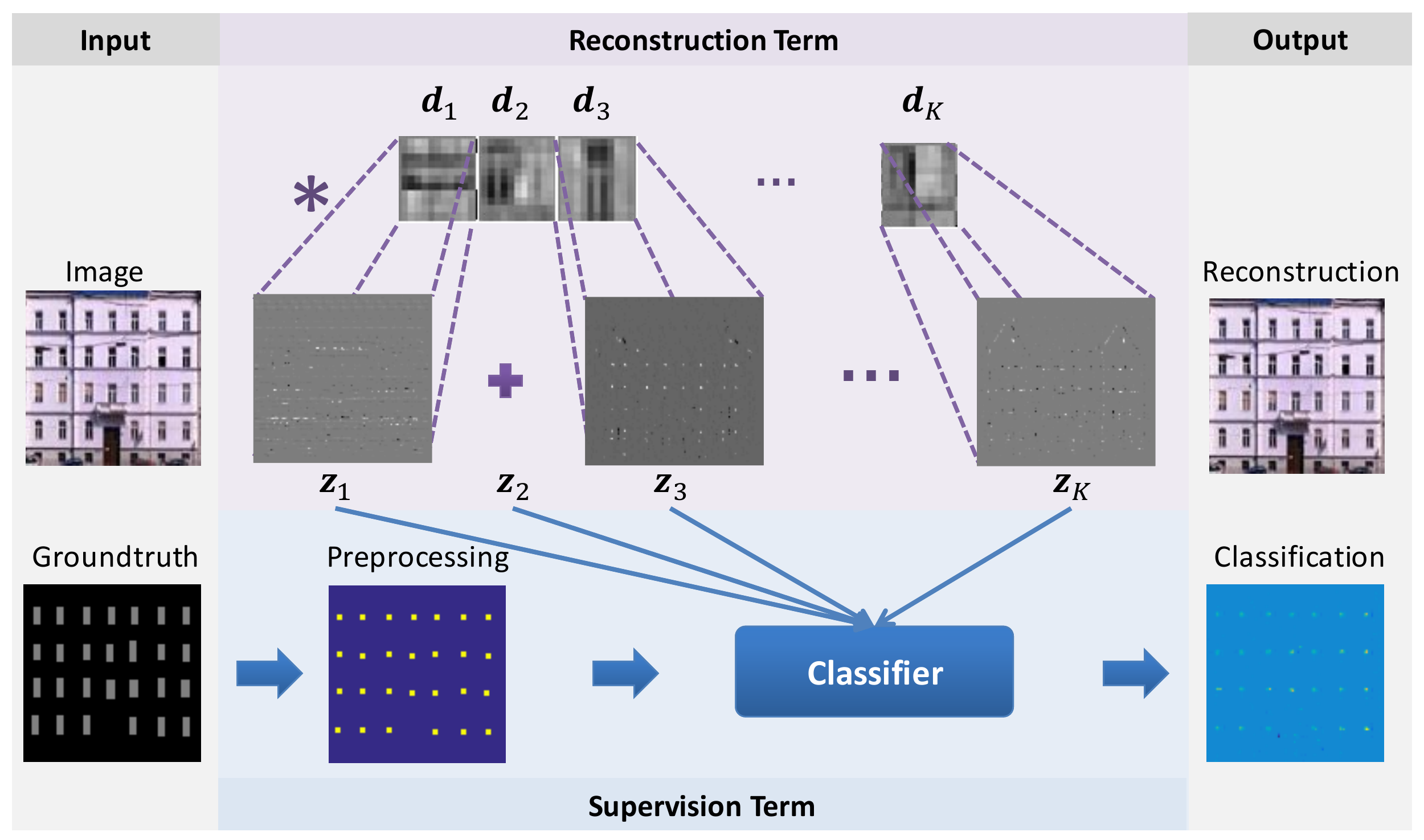}
	\centering
	\caption{Supervised Convolutional Sparse Coding Model. Top part shows traditional CSC where an input image is represented by a sum of dictionary elements convolved with sparse maps and resulting in a reconstructed image that is similar to the input image. An additional supervised term is added to aim the sparse maps towards producing positive labels at the center of groundtruth objects.}
	\label{fig:CSC_main}
\end{figure*}
The CSC problem can be expressed in the following form:
\begin{align}
\label{eq:CSC}
\begin{split}
\argmin_{\mathbf{d},\mathbf{z}}&\;\frac{1}{2}\|\mathbf{x}-\mathbf{M}\sum_{k=1}^K \mathbf{d}_k * \mathbf{z}_k \|_2^2 + \beta \sum_{k=1}^K \|\mathbf{z}_k\|_1\\
\text{subject to}&\;\; \|\mathbf{d}_k\|_2^2 \leq 1 \;\;\forall k \in \{1,...,K\}
\end{split}
\end{align}
where $\mathbf{d}_k \in \mathbb{R}^M$ are the vectorized 2D patches representing $K$ dictionary elements, $\mathbf{z}_k \in \mathbb{R}^D$ are the vectorized sparse maps corresponding to each of the dictionary elements, and $\mathbf{M}$ is a binary diagonal matrix for boundary handling (see Figure~\ref{fig:CSC_main}). The data term reconstructs the image $\mathbf{x} \in \mathbb{R}^D$ using a  sum of convolutions of the dictionary elements with the sparse maps, and $\beta$ controls the tradeoff between the sparsity of the feature maps and the reconstruction error. The inequality constraint on the dictionary elements assumes Laplacian distributed coefficients, which ensures solving the problem at a proper scale since a larger value of $\mathbf{d}_k$ would scale down the value of the corresponding $\mathbf{z}_k$ respectively. The above equation shows the objective function on a single image, and it can be easily extended to multiple images where for each image, $K$ sparse maps are inferred, whereas all the images share the same $K$ dictionary elements.

\subsection{CSC Subproblems}
The objective in Eq.~\ref{eq:CSC} is not jointly convex. However, solving it for a group of variables while keeping the others fixed leads to two convex subproblems, which we refer to as the coding subproblem and the dictionary learning subproblem. For ease of notation, we represent the convolution operations by multiplication of Toeplitz matrices with the corresponding variables.

\subsubsection{Learning Subproblem}
We learn the dictionary elements for a fixed set of sparse feature maps as shown in Eq.~\ref{eq:Learning}.
\begin{align}
\label{eq:Learning}
\begin{split}
\argmin_{\mathbf{d}}&\;\;\frac{1}{2}\|\mathbf{x}-\mathbf{MZ}\mathbf{S}^T\mathbf{d}\|_2^2\\
\text{subject to}&\;\; \|\mathbf{d}_k\|_2^2 \leq 1 \;\;\forall k \in \{1,...,K\}
\end{split}
\end{align}
Here, $\mathbf{Z}=[\mathbf{Z}_1\dots\mathbf{Z}_K]$ is of size $D \times DK$ and is a concatenation of the sparse convolution matrices, $\mathbf{d}=[\mathbf{d}_1^T\dots \mathbf{d}_K^T]^T$ is a concatenation of the dictionary elements, and $\mathbf{S}^T$ pads the filter to allow larger spatial support.\\

\subsubsection{Coding Subproblem}\label{codingsection}
We infer the sparse maps for a fixed set of dictionary elements as shown in Eq.~\ref{eq:Inference}.
\begin{equation}
\label{eq:Inference}
\argmin_{\mathbf{z}}\;\;\frac{1}{2} \|\mathbf{x}-\mathbf{MD}\mathbf{z}\|_2^2 + \beta\|\mathbf{z}\|_1\\
\end{equation}
Similar to above, $\mathbf{D}=[\mathbf{D}_1\dots\mathbf{D}_K]$ is of size $D \times DK$ and is a concatenation of the convolution matrices of the dictionary elements, and $\mathbf{z}=[\mathbf{z}_1^T\dots \mathbf{z}_K^T]^T$ is a concatenation of the vectorized sparse maps.\\

\subsection{CSC Optimization}\label{sec:csc_optimization}
Finding an efficient solution to the CSC problem is a challenging task due to its high computational complexity and the non-convexity of its objective function. Seminal advances~\cite{Bristow2013,Kong2014,Heide2015} in CSC have demonstrated computational speed-up by solving the problem efficiently in the Fourier domain where the convolution operator is transformed to element-wise multiplication. As such, the optimization is modeled as a biconvex problem with two convex subproblems that are solved iteratively and combined to form a coordinate descent solution. Despite the performance boost attained by solving the CSC optimization problem in the Fourier domain, the problem is still deemed computationally heavy due to the dominating cost of solving large linear systems. More recent work ~\cite{Heide2015,Kong2014,wohlberg2014efficient} makes use of the block-diagonal structure of the matrices involved and solves the linear systems in a parallel fashion.
To efficiently solve the subproblems, they are reformulated in~\cite{Heide2015} as sum of functions that are simple to optimize individually as such:
\begin{flalign*}
&\text{Coding Subproblem:}\;\;\;\argmin_{\mathbf{z}}\;\;f_1(\mathbf{Dz}) + f_2(\mathbf{z}) &\\
&\text{Learning Subproblem:}\;\argmin_{\mathbf{d}}\;\;f_1(\mathbf{Zd}) +  f_3(\mathbf{d}) &
\label{eq:sumfun_coding}
\end{flalign*}
where $$f_1(\mathbf{v})=\frac{1}{2}\|\mathbf{x}-\mathbf{Mv}\|_2^2,\;
f_2(\mathbf{v})=\beta\|\mathbf{v}\|_1,\;$$
$$f_3(\mathbf{v})=\sum_{k=1}^K ind_R(\mathbf{v}_k),\;
R=\{\mathbf{v}\;|\;\|\mathbf{Sv}\|_2^2 \leq 1\}$$

\noindent The subproblems shown above can be remapped to a general form that can be solved using existing optimization algorithms. ADMM in general solves equations of the general form shown below.
\begin{equation}
\label{eq:admm}
\argmin_{\mathbf{\mathbf{y,u}}}~~ h(\mathbf{y}) + g(\mathbf{u})\;\;
\text{s.t. }\mathbf{Ky}=\mathbf{u}
\end{equation}
We can map this general form to the coding and learning CSC subproblems by setting $h=0$ and $g(\mathbf{u})=\sum_i f_i(\mathbf{u})$, with $\mathbf{K}=[\mathbf{D}^T\;\mathbf{I}]^T,\;\mathbf{y}=\mathbf{z},\;i=\{1,2\}$ for the coding subproblem and $\mathbf{K}=[\mathbf{Z}^T\;\mathbf{I}]^T,\;\mathbf{y}=\mathbf{d},\;i=\{1,3\}$ for the learning subproblem.

Solving the subproblems using scaled ADMM leaves us with a minimization that is separable in all the $f_i$ as shown in Algorithm~\ref{alg:CSC}.
\begin{algorithm}
\begin{algorithmic}[1]
\While {not converged}
\State $\mathbf{y}^{k+1}=\argmin_{\mathbf{y}} \|\mathbf{Ky}-\mathbf{u}^k + \lambda^k\|_2^2$
\State $\mathbf{u}_i^{k+1}=\mathbf{prox}_{\frac{f_i}{\rho}}(\mathbf{K}_i\mathbf{y}_i^{k+1}+\mathbf{\lambda}_i^{k})$
\State $\mathbf{\lambda}^{k+1}=\mathbf{\lambda}^{k}+(\mathbf{Ky}^{k+1}-\mathbf{u}^{k+1})$
\EndWhile
\end{algorithmic}
\caption{ADMM for CSC Subproblems}
\label{alg:CSC}
\end{algorithm}
The second line of the algorithm involves a large linear system that can be  efficiently solved in the Fourier domain. The circulant convolution matrices $\mathbf{D}$ and $\mathbf{Z}$ become diagonal in the Fourier space and thus inverting them can be done in parallel by making use of the Woodbury formula for inverting  block diagonal matrices. The technical report by~\cite{Kong2014} gives more details to the solution. The third line constitutes proximal operators for each of the $f_i$s that are simple to derive and well known in the literature. In the paper by~\cite{Heide2015}, complexity and convergence details of this algorithm are shown.

In the next section, we show how to extend the formulation above to obtain a mathematical formulation for supervised CSC and we discuss the optimization algorithm used to compute an approximate solution.

\section{Supervised Convolutional Sparse Coding}
\label{sec:scsc}
In this work, given the input images, we consider that each pixel may belong to any of $p$ different classes coming from a variety of ground truth annotations including segmentation or bounding boxes. Now, we seek to derive a convolutional sparse representation that not only reconstructs the image, but also resembles the classification available in the annotation (see Figure~\ref{fig:CSC_main}). In this sense, the learned dictionary elements are guided to become supervised and more representative of the available classes.

\subsection{Supervised CSC Model}
The SCSC problem can be expressed in the following form:
\begin{align}
\label{eq:SCSC}
\begin{split}
\argmin_{\substack{\mathbf{d}_{1,...,K},\\\mathbf{z}_{1,...,K},\\\mathbf{\theta}=\{\mathbf{w},b\}}}&\;\frac{1}{2}\|\mathbf{x}-\mathbf{M}\sum_{k=1}^K \mathbf{d}_k * \mathbf{z}_k \|_2^2 + \beta \sum_{k=1}^K \|\mathbf{z}_k\|_1+ \gamma(C(\mathbf{y},\mathbf{z}_{1...K},\theta)+\alpha\|\theta\|_2^2)\\
\text{subject to}&\;\; \|\mathbf{d}_k\|_2^2 \leq 1 \;\;\forall k \in \{1,...,K\}\\
\end{split}
\end{align}
Here, we take  $C$  to be the logistic regression function: $C=\sum_{d=1}^{\tilde{D}} \log(1+e^{-y_d(\mathbf{w}^T\mathbf{z}_d+b)})$, where $y_d\in\{-1,+1\}$ constitutes the associated pixelwise groundtruth labels, $\mathbf{\theta}$ parameterizes the classification model with the linear classification coefficients $\mathbf{w}$ and $b$, and $\tilde{D}<D$ is the number of selected groundtruth pixels. $\alpha$ is the regularization parameter that prevents overfitting and $\gamma$ is the tradeoff parameter between reconstruction and classification. The above formulation shows the objective for the case of two classes and it can be easily extended into multiple classes in a one vs. all framework. This is similar to the approach of~\cite{mairal2009supervised} for supervised dictionary learning.\\

\subsection{Supervised CSC Subproblems.}
The addition of a logistic regression loss function as a regularizer to the CSC problem gives rise to three subproblems with an additional subproblem for optimizing the weights of the linear classification model. We will refer to the third subproblem as the classification subproblem. As in unsupervised CSC, the overall objective is not convex, but it is convex when holding all variables  except  one fixed.\\

\noindent \textbf{Learning Subproblem.}~We will start with the learning subproblem where we need to optimize the dictionary elements $\mathbf{d}_{1,...,K}$ given the sparse codes and the classifier parameters. Here, we end up with a subproblem that is exactly the same as in unsupervised CSC shown in Equation~\ref{eq:Learning}, since the supervision regularizer is independend of the dictionary.\\

\noindent \textbf{Supervised Coding Subproblem.}~In the coding subproblem, we infer the sparse maps $\mathbf{z}_{1,...,K}$ given the dictionary and the classifier parameters as shown in Equation~\ref{eq:coding_supervised}.
\begin{align}
\begin{split}
\label{eq:coding_supervised}
\argmin_{\mathbf{z}}\;\;&\frac{1}{2} \|\mathbf{x}-\mathbf{MD}\mathbf{z}\|_2^2 + \beta\|\mathbf{z}\|_1+ \gamma \sum_d^{\tilde{D}} \log(1+e^{-y_d(\mathbf{\tilde{w}}^T_d\mathbf{z}+b)})
\end{split}
\end{align}
Here, $\mathbf{\tilde{w}}^T_d$ corresponds to the $d^{th}$ row of the matrix $\mathbf{W}=[w_1\mathbf{I}\;w_2\mathbf{I}\;...\;w_K\mathbf{I}]$ that is a concatenation of diagonal matrices, each one representing the support for the features taken from the sparse maps. The sparse maps here are added to the classification term which serves as an additional regularizer to the problem. Unlike the unsupervised coding subproblem in section~\ref{codingsection},  the reconstruction is not the only factor involved in computing the sparse maps. A good sparse map in the supervised coding task would trade-off some reconstruction to become more discriminative.\\

\noindent \textbf{Classification Subproblem.}~The classification subproblem is a  regular logistic regression problem as shown in Equation~\ref{eq:regression}. It can be solved  using gradient descent. 
\begin{equation}
\label{eq:regression}
\argmin_{\mathbf{w},b}\;\;\sum_{d=1}^{\tilde{D}} \log(1+e^{-y_d(\mathbf{w}^T\mathbf{z}_d+b)})+\alpha(\|\mathbf{w}\|_2^2+b^2)
\end{equation}

\subsection{Supervised CSC Optimization}
We will follow the approach presented in~\cite{Heide2015} for solving the supervised CSC problem. We will represent the learning and coding subproblems as a sum of simple convex functions. With this approach, the learning subproblem can be solved exactly, similar to unsupervised CSC, as shown in Section~\ref{sec:csc_optimization}. \\
The coding subproblem has an additional function holding the logistic function fitting term, and thus can be reformulated as such:
\begin{align}
\begin{split}
\argmin_{\mathbf{z}}\;\;f_1(\mathbf{Dz}) + f_2(\mathbf{z}) +f_4(\mathbf{Wz})\\
\text{where }f_4(\mathbf{v})=\gamma \sum_d^{\tilde{D}} \log(1+e^{-y_d(v_d+b)})
\end{split}
\end{align}
This formulation allows us to follow the same solution by casting the subproblem into the general form of ADMM shown in Equation~\ref{eq:admm} with a minor modification of the variables $\mathbf{K}=[\mathbf{D}^T\;\mathbf{I}\;\mathbf{W}^T]^T$ and $i=\{1,2,4\}$. \\
The solution of the coding subproblem follows the steps presented in Algorithm~\ref{alg:CSC}. The second line involves solving a quadratic least squares problem with the solution:
\begin{equation}
\mathbf{z}_{\text{opt}}=(\mathbf{K}^T\mathbf{K})^{-1}(\mathbf{K}^T(\mathbf{u}-\mathbf{\lambda}))
\end{equation}
The inverse can be solved efficiently in the Fourier domain since similar to $\mathbf{D}$, $\mathbf{W}$ is a concatenation of diagonal matrices and one can find a variable reordering that makes $(\mathbf{D}^\dagger\mathbf{D}+\mathbf{I}+\mathbf{W}^\dagger\mathbf{W})^{-1}$ block diagonal, thus making the inversion parallelizable over the $D$ blocks. Using the Woodburry formula, the inverse can be computed as:
\begin{equation}
(\mathbf{D_j}^\dagger\mathbf{D_j}+\tilde{\mathbf{W}})^{-1}=\tilde{\mathbf{W}}^{-1}
-\frac{\tilde{\mathbf{W}}^{-1}\mathbf{D}_j^\dagger\mathbf{D}_j\tilde{\mathbf{W}}^{-1}}{1+\mathbf{D}_j\tilde{\mathbf{W}}^{-1}\mathbf{D}_j^\dagger}
\end{equation}
where $\tilde{\mathbf{W}}=\mathbf{w}\mathbf{w}^T+\mathbf{I}$ is the same over all the blocks whose inverse can be computed beforehand.\\
In addition, the third line of the algorithm for supervised CSC includes an additional proximal operator for the logistic loss function. Setting its gradient to zero constitutes solving an intersection of an exponential with a line. Unfortunately, no closed form solution to this operator exists in the literature but it can be efficiently computed using Newton's method.

\begin{algorithm}
\begin{algorithmic}[1]
\State Initialize $\mathbf{d},\mathbf{\theta},\mathbf{\lambda}_{d,z}$
\While {not converged}
\State Update $\mathbf{z}$: Solve Eq.~\ref{eq:coding_supervised} using Alg.~\ref{alg:CSC}
\State Update $\mathbf{d}$: Solve Eq.~\ref{eq:Learning} using Alg.~\ref{alg:CSC}
\State Update $\mathbf{\theta}$: Solve Eq.~\ref{eq:regression} using Gradient Descent
\EndWhile
\end{algorithmic}
\caption{Coordinate Descent for SCSC}
\label{alg:SCSC}
\end{algorithm} 

\section{Experimental Evaluation}\label{sec:results}
In this section, we show results for applying supervised CSC (denoted as SCSC) presented in Algorithm~\ref{alg:SCSC} compared with unsupervised CSC as a baseline approach using the implementation of~\cite{Heide2015}.
We describe implementation details, the quantitative and qualitative evaluation of the discriminability of the learned dictionaries, the reconstructive quality of the disriminative dictionaries, and inpainting results.

\subsection{Implementation Details}

We tested our algorithm on different datasets: \emph{Graz}~\cite{graz} made of 50 images and having groundtruth labels for the window class, \emph{ecp}~\cite{ecp} made of 104 images using window and balcony labels,  \emph{label-me}~\cite{labelme} made of a random subset of 50 images using the tree and building classes, and \emph{coco}~\cite{coco} on a random subset of 50 images using the cat class. In most experiments, we used 10 images for training on \emph{Graz}, \emph{label-me} and \emph{coco} datasets and 25 images for training on \emph{ecp}.

For \emph{Graz} and \emph{ecp}, since we need to target the sparse maps to fire at the center of windows and thus learn dictionary elements that are similar to window patches, we apply a preprocessing step where we restrict the positive samples to the centers of the windows as shown in Figure~\ref{fig:image}-(d). For \emph{coco} and \emph{label-me}, due to the high variation in scale and appearance of the groundtruth segmentations in this dataset, we do not apply a preprocessing step and thus seek dictionary elements that are representative of sub-patches of the entire groundtruth segmentation. As shown in Figure~\ref{fig:image}-(d), there exists class imbalance because background pixels are dominant. Thus, we make sure to sample a similar number of positive and negative samples to make the classification less affected by this imbalance.

The scale at which the dictionaries are learned greatly affects the quality of the learning and classification. In our experiments, we learn dictionaries of size $11\times11$ since objects in our datasets are generally at this scale or smaller. We set the sparsity coefficient $\beta=0.5$ and the logistic regression regularization parameter $\alpha=1$. In the following experiments, we use the average precision (AP) score of each class to evaluate  the classification accuracy and the peak signal to noise ratio (PSNR) to evaluate the quality of  reconstruction.

\subsection{Discriminative Dictionary Results}

\begin{figure}[t]
	\centering
		\includegraphics[width=0.9\linewidth]{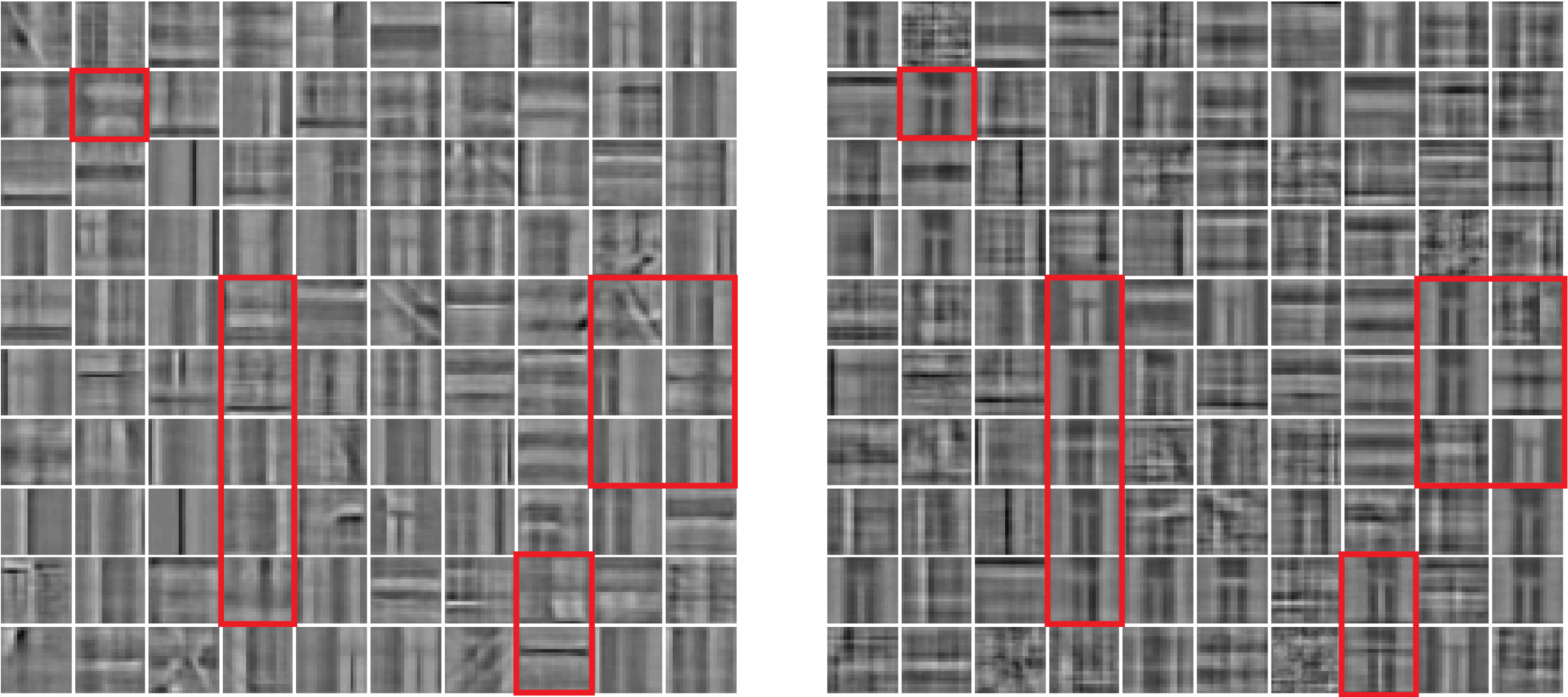}
	\centering
	\caption{Learned elements using unsupervised (left) and supervised CSC (right).}
	\label{fig:comparedics}
\end{figure}

We first show the learned filters converged by our Supervised CSC optimization algorithm. Starting from random initial filters, SCSC proceeds iteratively transforming random pixels to more meaningful filters reflecting the structure of the corresponding class. To demonstrate how the supervised dictionaries compare to unsupervised ones, we show in Figure~\ref{fig:comparedics} $K=100$ learned $21\times21$ filters from the \emph{Graz} dataset. By visual inspection, the supervised dictionaries show patches that are more representative of the classification task at hand. Although the two problems are initialized from the same random patches, supervised CSC converges to dictionary elements that resemble the window class on which we perform the supervision (see elements highlighted in red). The accompanying video in \textbf{supplementary material} also shows the progression of the learned filters during the optimization.\\

\begin{figure}[t]
	\centering
		\includegraphics[width=0.8\linewidth]{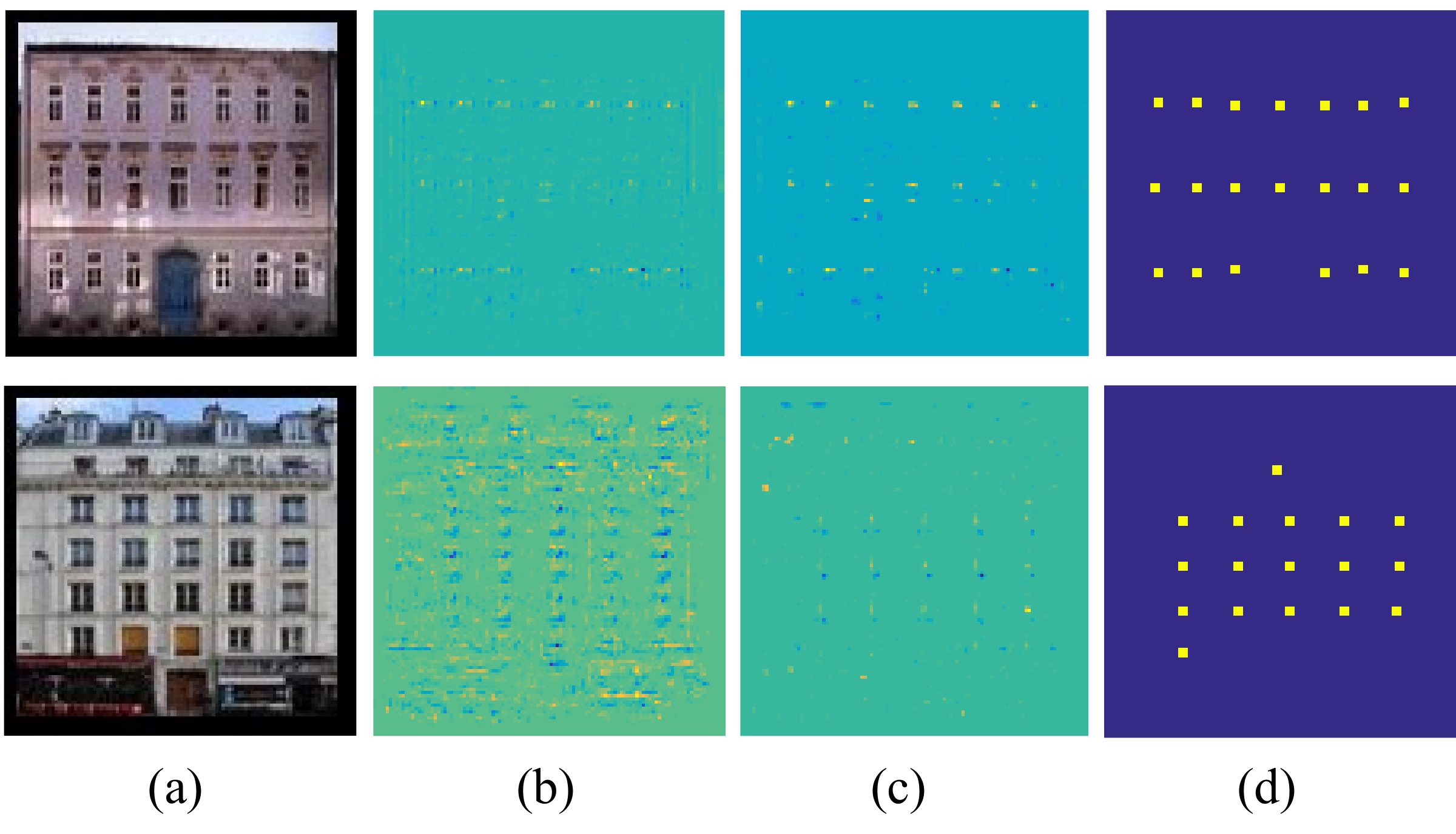}
	\centering
	\caption{(a)Input image, (b) unsupervised CSC classification result, (c) supervised CSC classification result, and (d) groundtruth labels.}
	\label{fig:image}
\end{figure}
\begin{figure*}[t]
	\subfloat[][]{
        \includegraphics[width=0.32\linewidth]{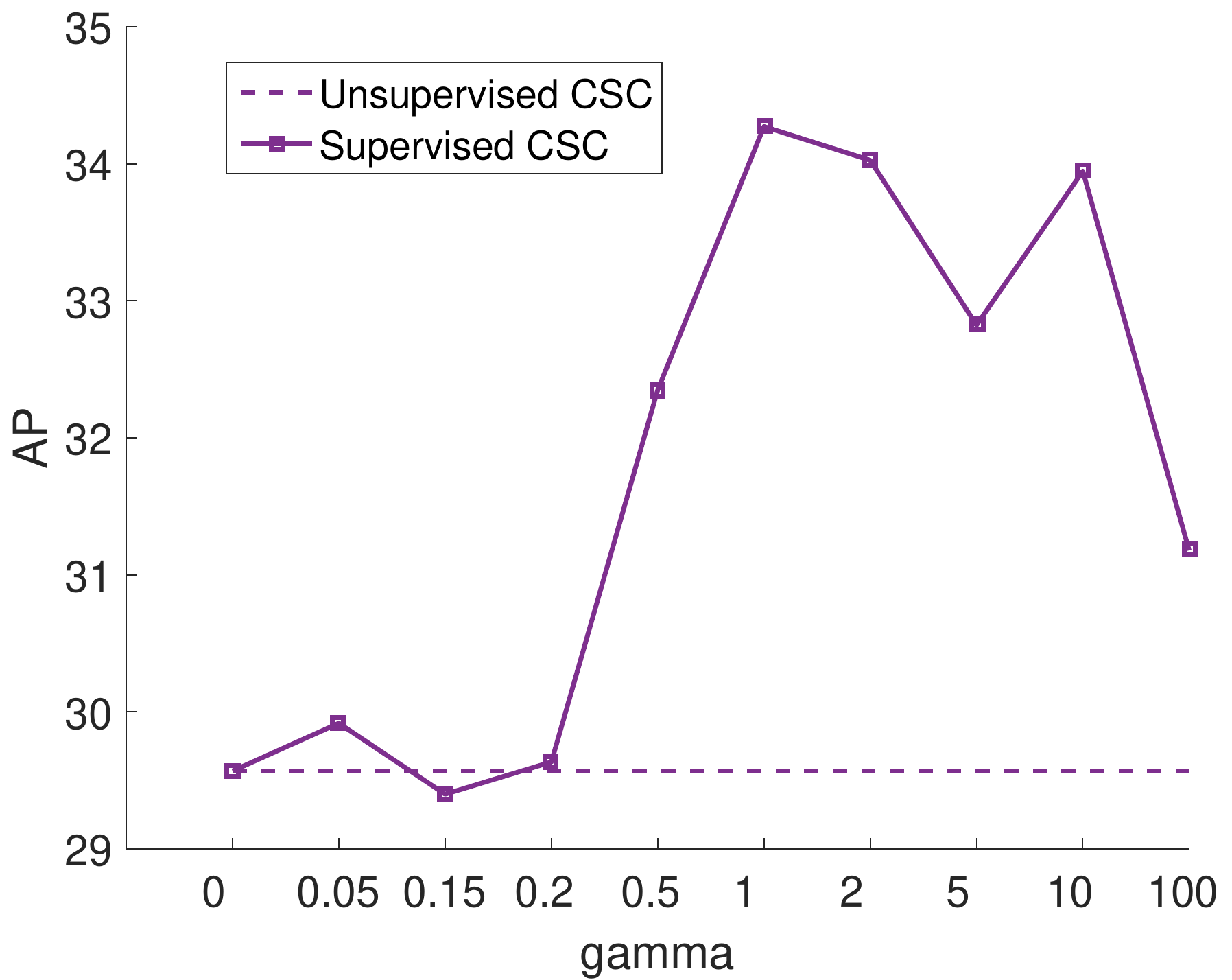}
        \label{fig:ap_gamma}
    }
    \subfloat[][]{
        \includegraphics[width=0.32\linewidth]{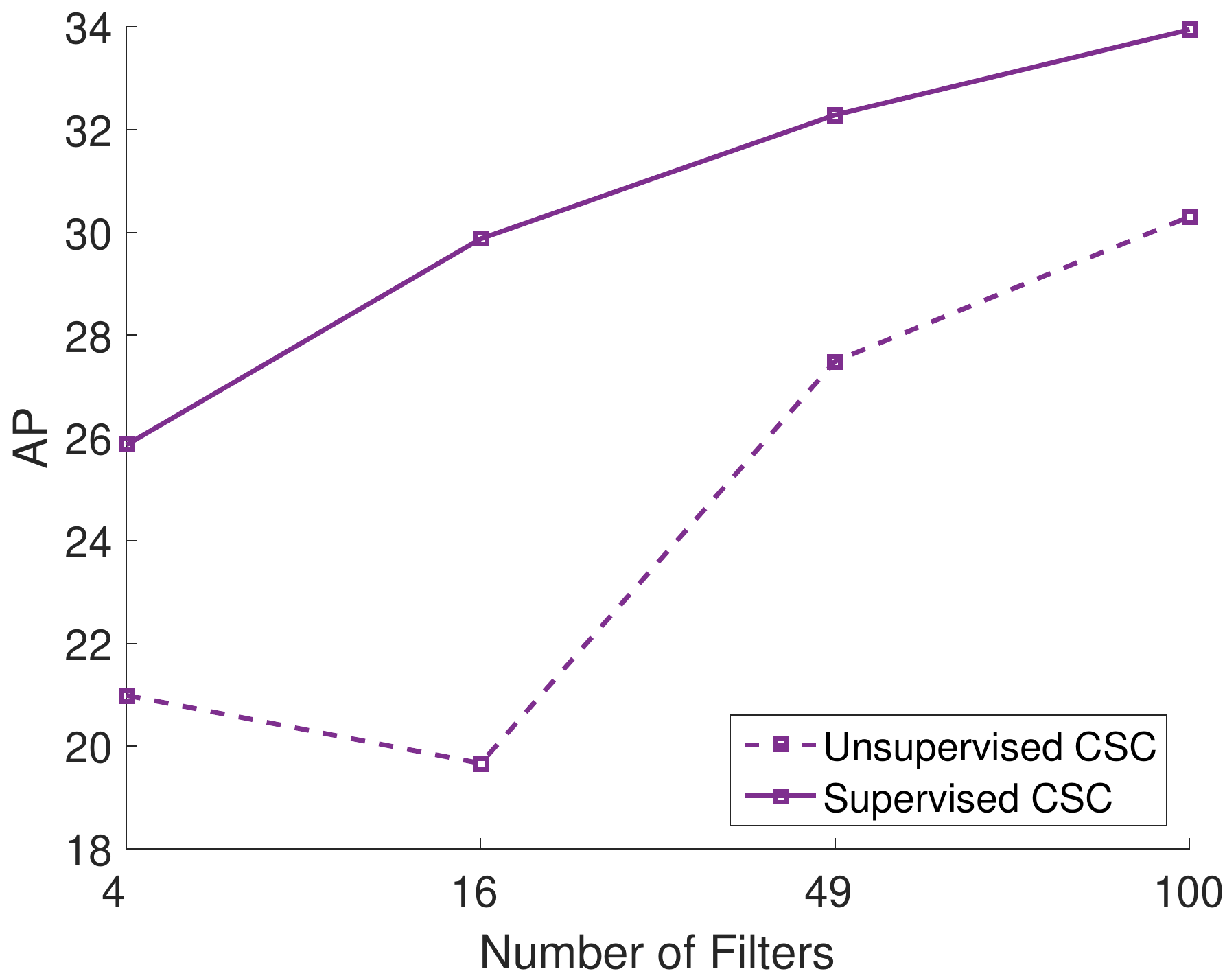}
	    \label{fig:ap_k}
    }
    \subfloat[][]{
        \includegraphics[width=0.32\linewidth]{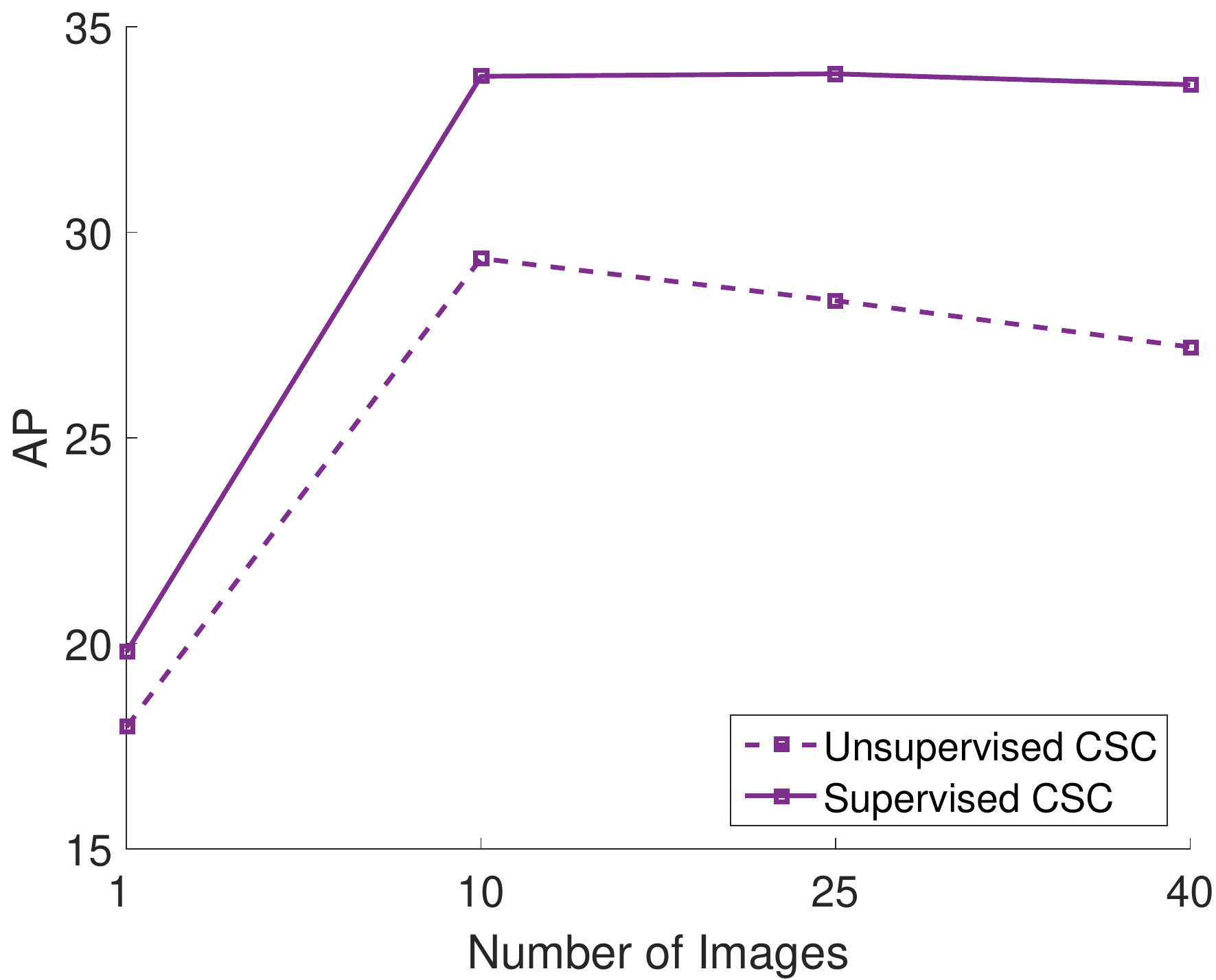}
	    \label{fig:ap_n}
    }

    \subfloat[][]{
        \includegraphics[width=0.32\linewidth]{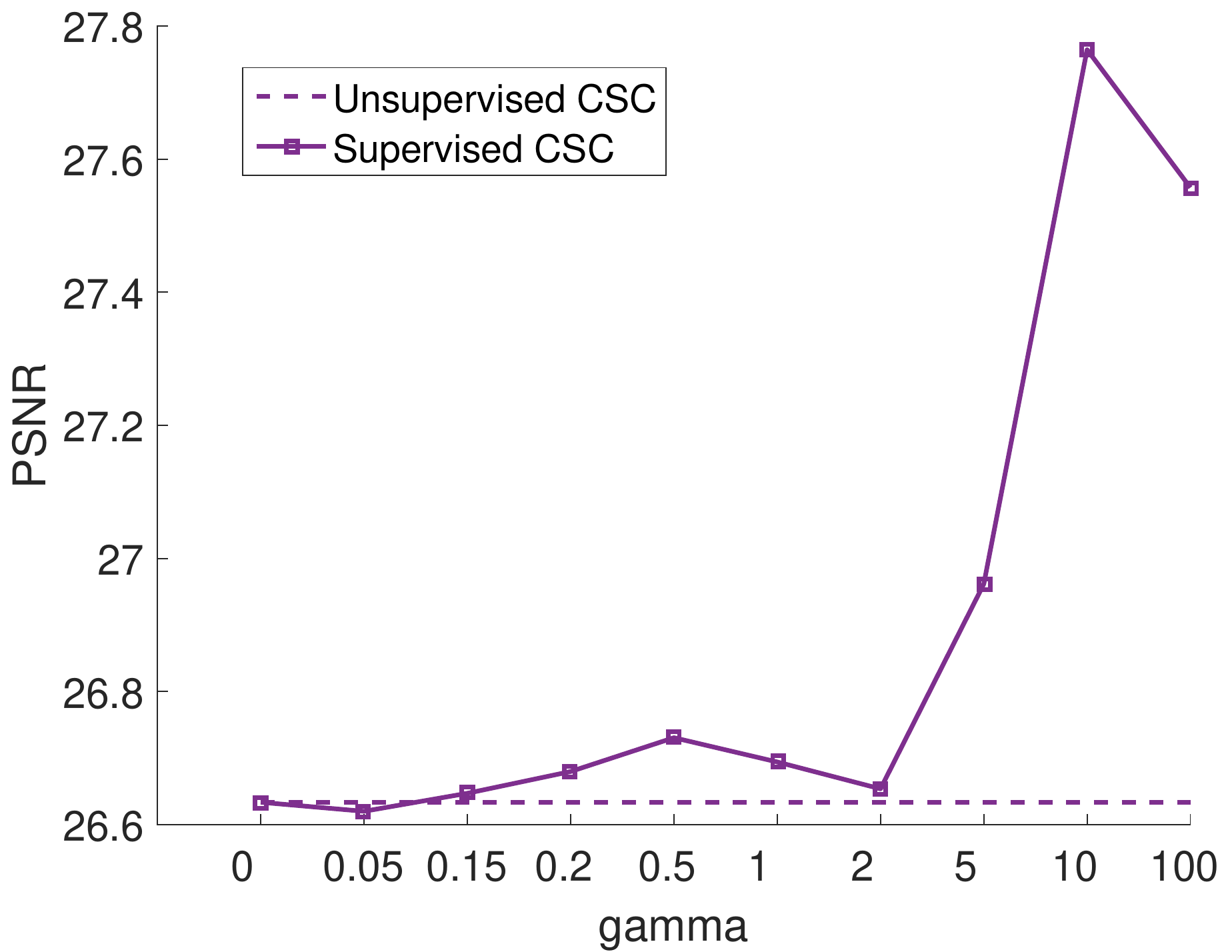}
	    \label{fig:psnr_gamma}
    }
    \subfloat[][]{
        \includegraphics[width=0.32\linewidth]{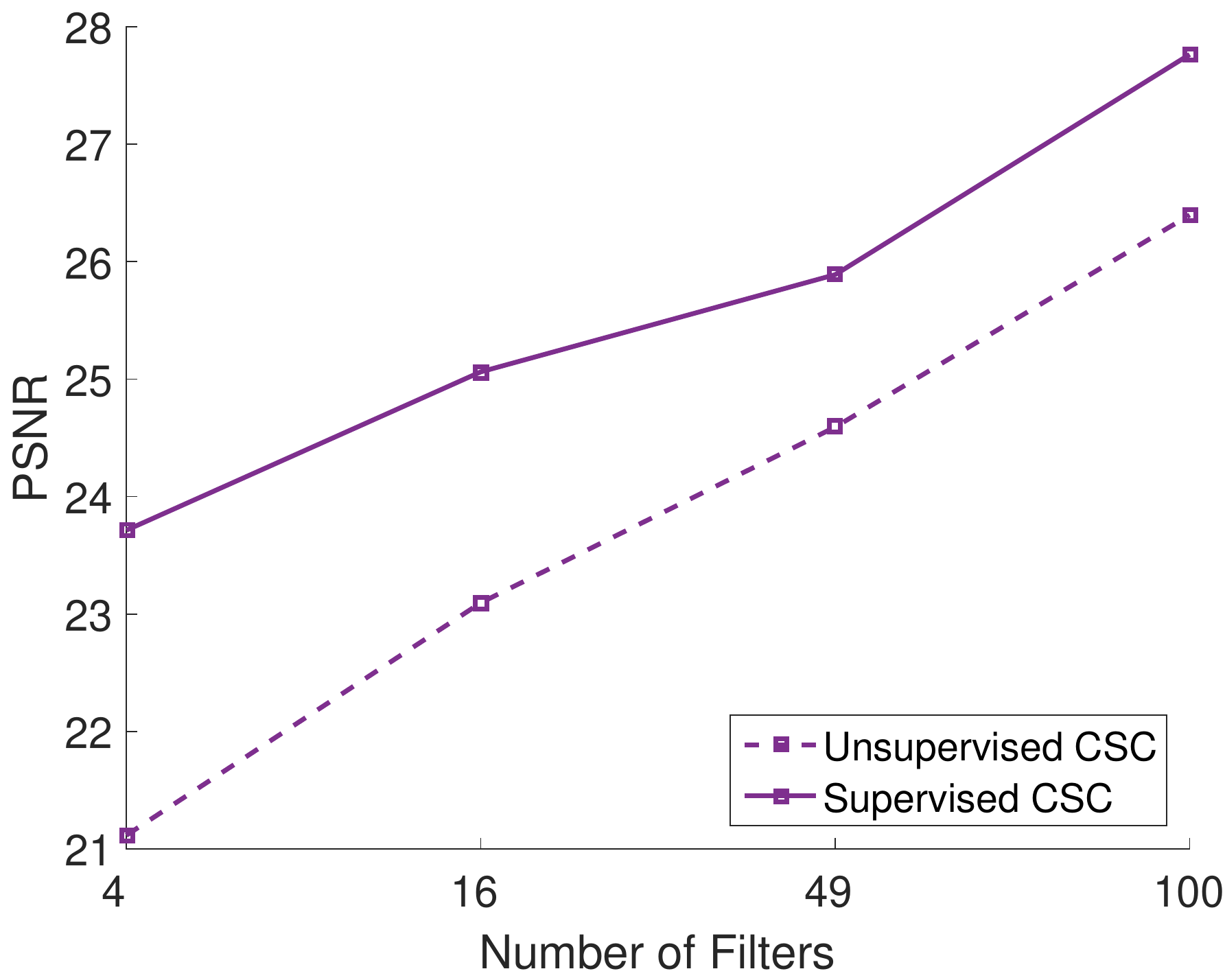}
	    \label{fig:psnr_k}
    }
    \subfloat[][]{
        \includegraphics[width=0.32\linewidth]{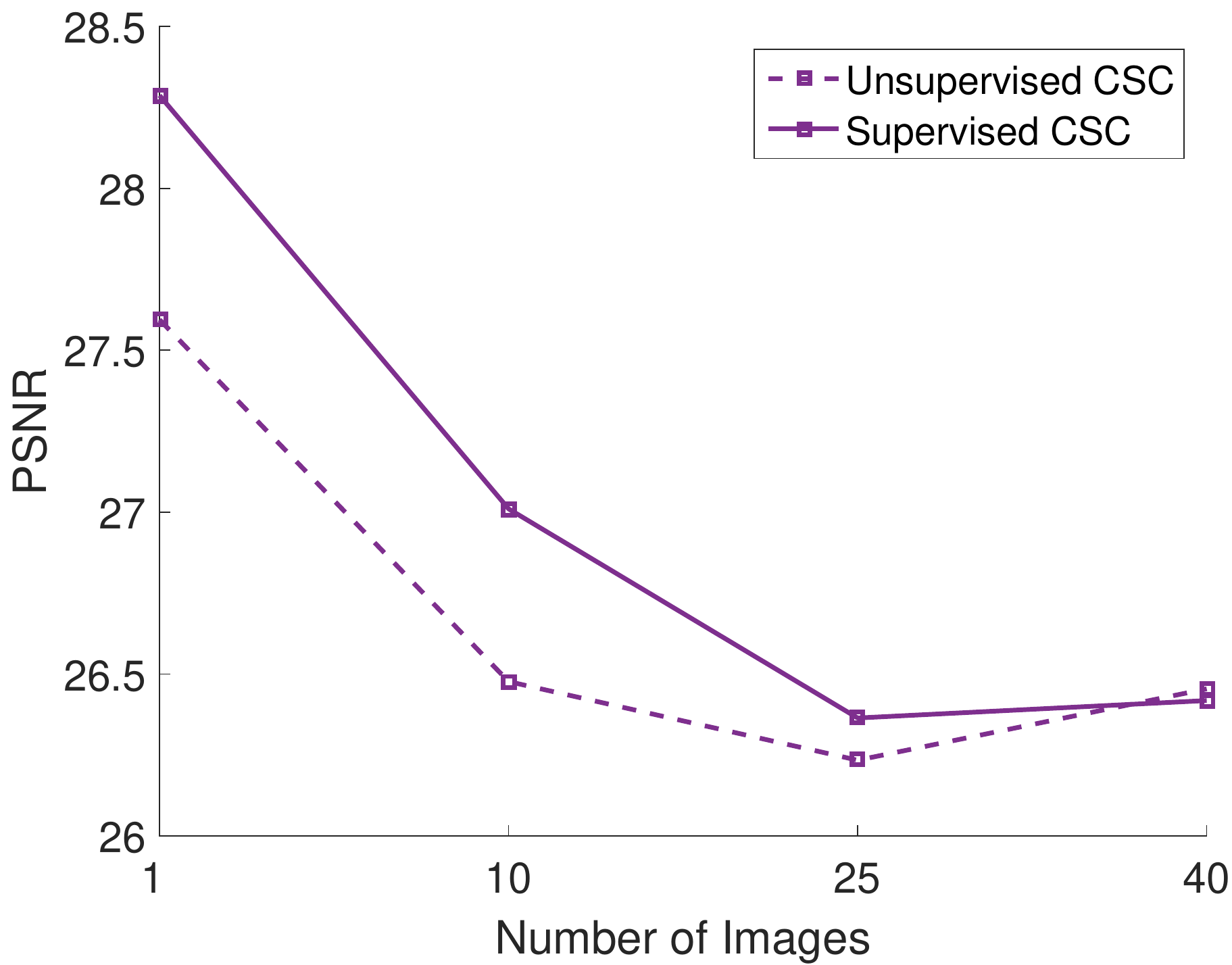}
	    \label{fig:psnr_n}
    }
	\caption{(a) AP score and (d) PSNR score on \emph{Graz} dataset as the classification tradeoff $\gamma$ increases. (b) AP score and (e) PSNR score as the number of filters $K$ increases. (c) AP score and (f) PSNR score as the number of images $N$ increases.}
	\label{fig:TimeIterK}
\end{figure*}

In addition, to verify that the learned dictionaries have semantic meaning, we show classification results on two example images. Please note that our goal is not to achieve state of the art image classification results, but only to show that the learned filters have semantic meaning, i.e. are more discriminative.
Figure~\ref{fig:image}-(b) and (c) show the label predictions (estimated by the learned logistic regression model trained on the sparse maps as features) corresponding to the unsupervised and supervised CSC approaches respectively. We do not directly use the classifier parameters inferred from the SCSC dictionary learning phase. A simple logistic regression classifier is retrained on the sparse maps inferred from unsupervised vs. supervised dictionaries and used to generate the predictions for the image pixels. As shown in th e figure, SCSC class predictions are much more distinctive as compared with unsupervised CSC. Background pixels get oppressed and the structure of the windows gets more prevalent when using supervised dictionaries.\\
We also show how the learnt dictionaries boost the performance of classification on unseen test sets. We first vary the classification coefficient $\gamma$ to show the effect of giving higher weight for the classification loss. Figure~\ref{fig:ap_gamma} shows the AP score as $\gamma$ increases for the \emph{Graz} dataset. In addition, Table~\ref{table} shows the AP score for the window and balcony classes in the \emph{ecp} dataset. The increase in $\gamma$ generally increases the classification accuracy until it reaches a point where the AP drops. The drop happens since the reconstruction term becomes negligible for higher value of $\gamma$, which results in over-fitting of the dictionary elements to the training data without taking their appearance into account. We also show how the AP changes with varied number of filters $K$ and number of training images $N$. Figure~\ref{fig:ap_k} and~\ref{fig:ap_n} show that as the number of filters and images increases, the classification precision for SCSC is higher than that of CSC.

\subsection{Reconstruction Results}
\begin{figure*}[t]
	\subfloat[][]{
        \includegraphics[width=0.32\linewidth]{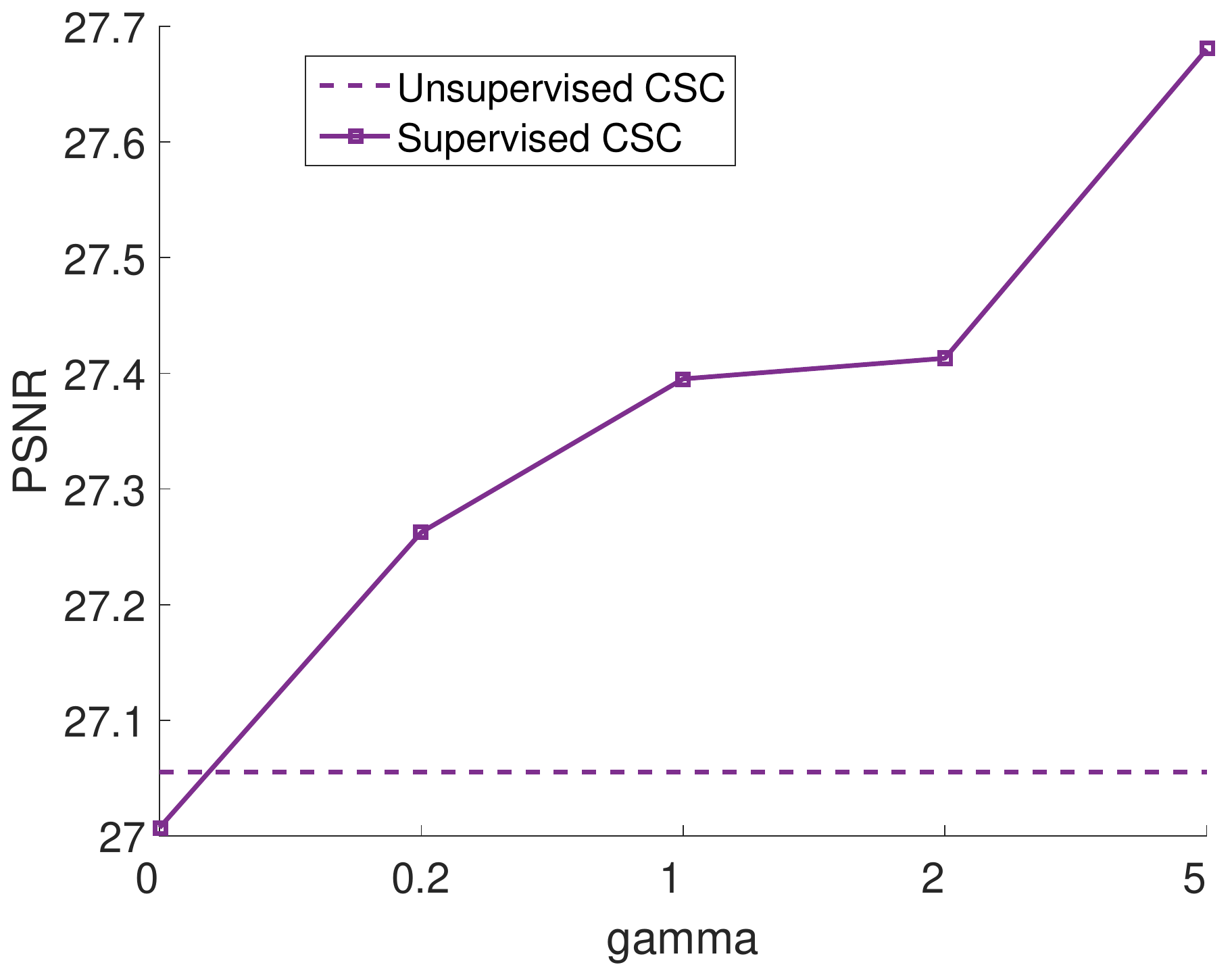}
        \label{fig:psnr_cat}
    }
    \subfloat[][]{
        \includegraphics[width=0.32\linewidth]{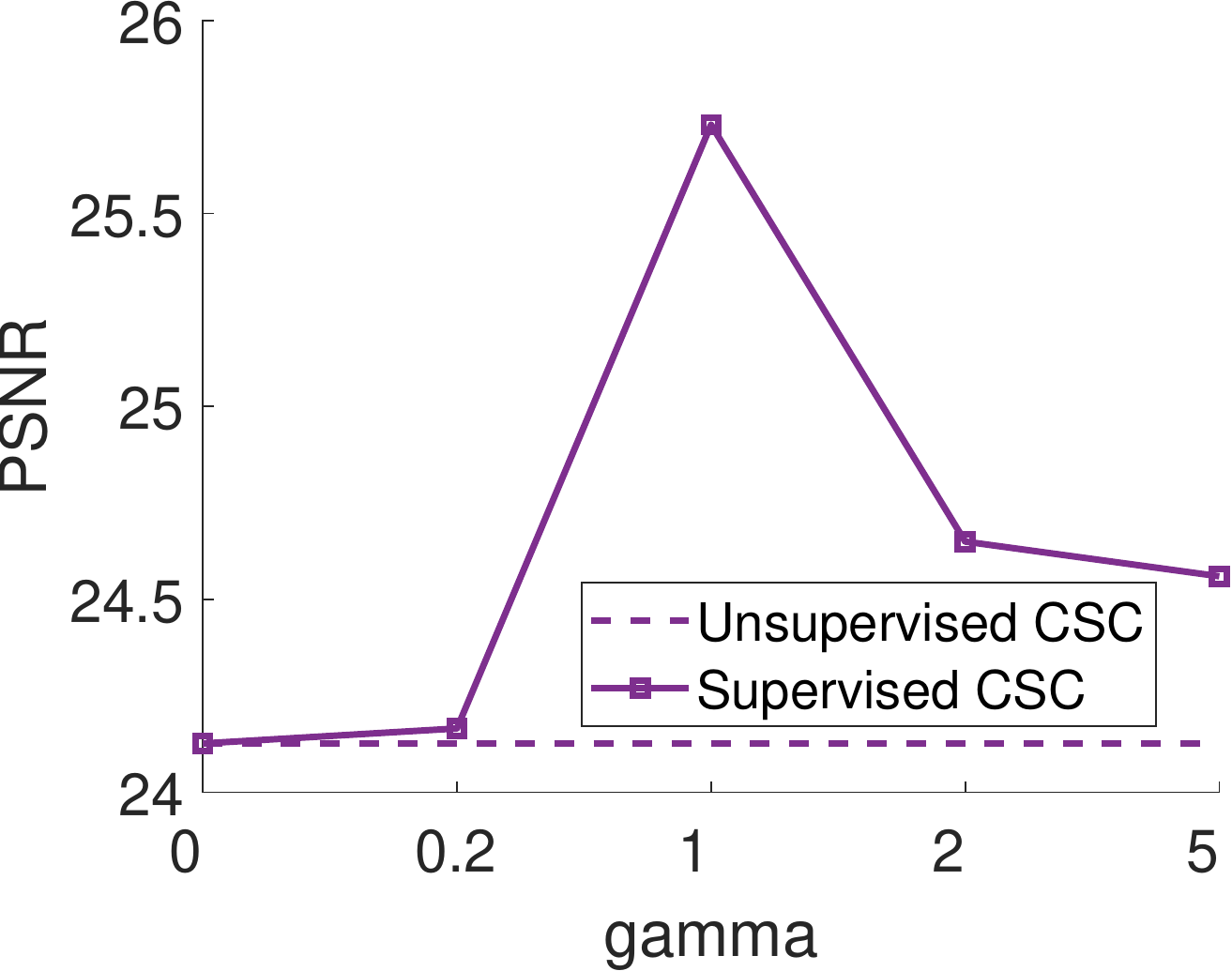}
	    \label{fig:psnr_building}
    }
    \subfloat[][]{
        \includegraphics[width=0.32\linewidth]{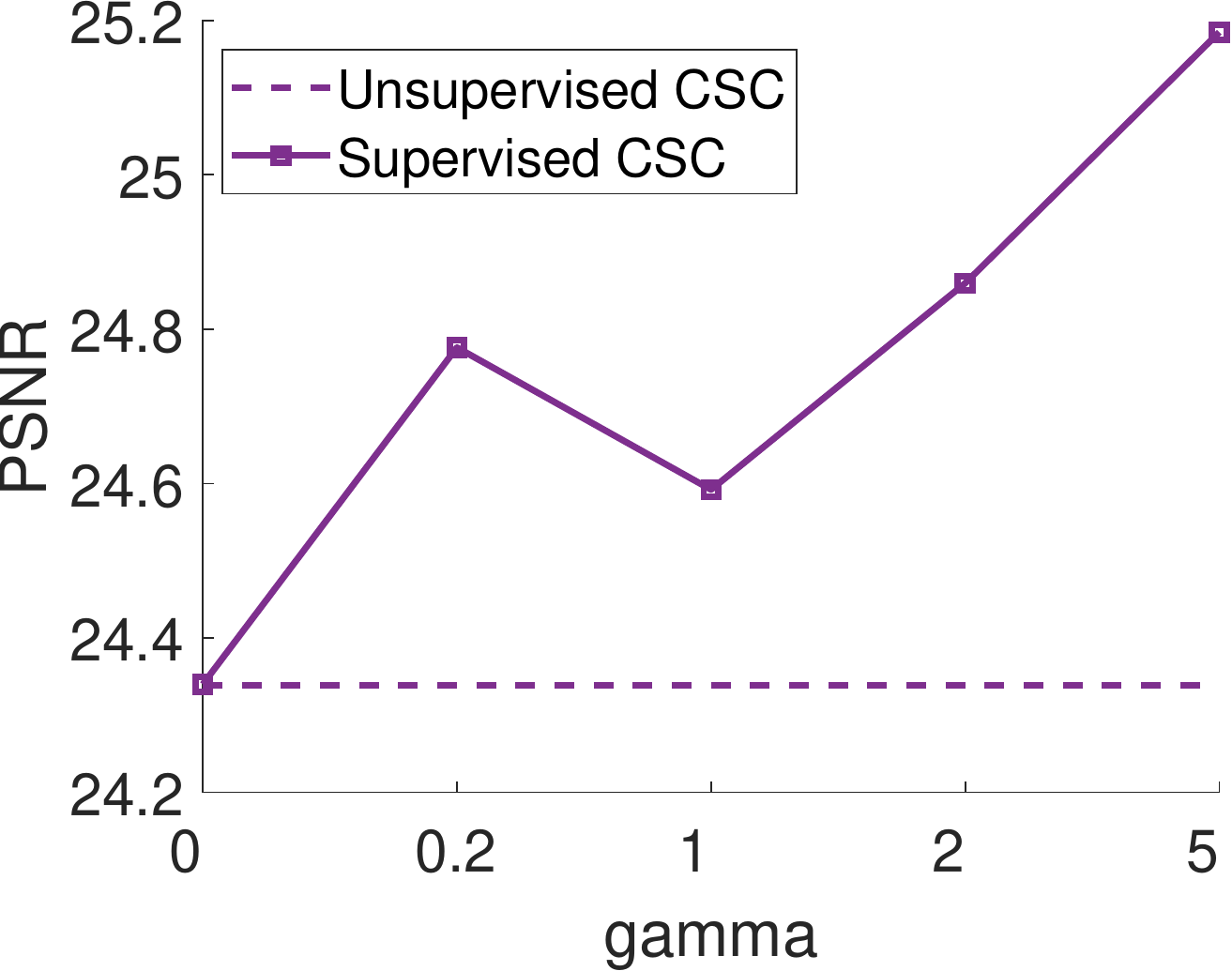}
	    \label{fig:psnr_tree}
    }

	\caption{PSNR score on (a) the cat class in \emph{coco} dataset, (b) building class and (c) tree class in \emph{label-me} dataset as the classification tradeoff $\gamma$ increases.}
		\label{fig:psnr_datasets}
\end{figure*}

In the following experiments, we validate that learning supervised dictionaries also improve the reconstruction quality on unseen images. The reconstruction quality is evaluated by the PSNR score where a higher score indicates a better reconstruction of the original image.\\
Figure~\ref{fig:psnr_gamma} and the last row in Table~\ref{table} show the positive effect of using supervised dictionary learning on the quality of the reconstruction where for some values of $\gamma$ the supervised approach achieves significantly higher PSNR values for the \emph{Graz} dataset and slightly improved values for the \emph{ecp} dataset. When the dictionary elements are discriminative enough they can generalize to different variations of the object class instead of overfitting to those appearing in the training set, which is what happens in unsupervised CSC.\\
We also show how the PSNR changes when the number of filters $K$ and number of training images $N$ vary. The number of filters corresponds to the number classifier features  and thus increasing $K$ gives a richer representation of the pixels. Figure~\ref{fig:psnr_gamma} shows that as the number of filters increases, the reconstruction quality increase for both CSC and SCSC with SCSC giving better reconstructions.\\
\indent To further verify the effect of reconstruction improvement for supervised CSC, we test our approach on classes from well known segmentation datasets. Figure~\ref{fig:psnr_datasets} shows that using supervised dictionaries, the reconstruction quality of images generally increases. Steering the dictionary elements to semantically meaningful representations allows learning filters that better reconstruct other instances of images within the supervised class.
\begin{table*}[t]
\centering
\caption{AP and PSNR of CSC and SCSC for varied $\gamma$ on the \emph{ecp} dataset classes.}
\label{table}
\begin{tabular}{l|c|c|c|c|c|c|c|c|}
\cline{2-9}
                                          & \textbf{CSC} & \multicolumn{7}{c|}{\textbf{Supervised CSC}}                                        \\ \hline
\multicolumn{1}{|l|}{\textbf{$\gamma$}}      & 0            & 0.2      & 0.5      & 1        & 2        & 5                 & 10       & 100      \\ \hline
\multicolumn{1}{|l|}{\textbf{Window AP}}  & 21.8     & 21.1 & 21.7 & 21.3 & 22.6 & \textbf{24.1} & 24 & 15.8 \\ \hline
\multicolumn{1}{|l|}{\textbf{Blacony AP}} & 4.6     & 7.2 & 7.3 & 8.2 & 11.7     & \textbf{14.1}     & 13.3     & 7        \\ \hline
\multicolumn{1}{|l|}{\textbf{PSNR}}       & 26.1         & 26.1     & 26.1     & 26.2     & 26.1     & \textbf{26.4}     & 26.3     & 26.1     \\ \hline
\end{tabular}
\end{table*}

\subsection{Image Inpainting Results}
The inclusion of the $\mathbf{M}$ matrix in the CSC optimization allows reconstructing incomplete images as discussed in~\cite{Heide2015}. Thus, we validate the performance of inpainting unseen images from the same dataset that we used for supervised learning vs. images from a different dataset. For inpainting, we randomly set $50\%$ of the pixel values to zero and  code the image with incomplete data. Table~\ref{inpaint} shows the reconstruction quality when comparing the reconstructed image with its original complete image. As shown, the inpainting results using supervised dictionaries generally lead to a better reconstruction quality. This is more evident on instances of images within the same supervised class where the PSNR value shows a higher boost compared to instances from another class where CSC and SCSC have comparable performance. This shows that the learned supervised dictionaries have semantic value that serves in better reconstructing general image instances from the supervised class unlike traditional CSC which learns dictionaries that overfits the training data.\\
\begin{table}[t]
\centering
\caption{PSNR score for image inpainting}
\label{inpaint}
\begin{tabular}{l|l|l|l|l|l|l|l|l|l|l|}
\cline{2-11}
                           & \multicolumn{10}{c|}{same dataset}                                                                                                                          \\ \hline
\multicolumn{1}{|l|}{CSC}  & 22.7          & 24.1          & 20.6          & 21            & \textbf{19.8} & 23.3          & 21.7        & \textbf{24.7} & 21.5          & 24            \\ \hline
\multicolumn{1}{|l|}{SCSC} & \textbf{23.5} & \textbf{25.1} & \textbf{22.6} & \textbf{21.5} & 17.7          & \textbf{24.7} & \textbf{24} & 24.2          & \textbf{22.2} & \textbf{24.2} \\ \hline
                           & \multicolumn{10}{c|}{other dataset}                                                                                                                         \\ \hline
\multicolumn{1}{|l|}{CSC}  & 21.8          & 20.5          & 19.6          & 18            & 18.3          & 21            & 19.5        & 18.6          & 19.5          & 17.7          \\ \hline
\multicolumn{1}{|l|}{SCSC} & 21.8          & 20.5          & \textbf{19.8} & \textbf{18.2} & \textbf{18.4} & \textbf{21.2} & 19.5        & \textbf{18.7} & \textbf{19.6} & \textbf{17.8} \\ \hline
\end{tabular}
\end{table}
The above results verify that using supervised convolutional sparse coding gives rise to learning shift invariant dictionaries that are not only very representative in reconstructing images but more semantically relevant than those trained in an unsupervised manner. While it is expected that our approach improves the classification performance, it is somewhat surprising that we can also improve the quality of the reconstruction. We attribute this to the fact that we make the filters semantically more meaningful and implicitly steer the optimization in the right direction to improve reconstruction performance. Embedding semantics into the dictionary makes it more generalizable to unseen images in which the objects appear with appearances that vary from what is seen in training data. This is opposed to unsupervised CSC which tends to overfit to the appearance of objects in its training set.

\section{Conclusion and Future Work}
In this work, we proposed a model for supervised convolutional sparse coding (SCSC) in which the CSC problem can be solved jointly for learning reconstructive dictionary elements while targeting a classification model. Results on multiple datasets showed more semantically meaningful, i.e. discriminative, filters when compared to regular CSC while at the same time improving the reconstruction quality. We believe that this is a very surprising and interesting result that our more discriminative filters are also better suitable for reconstruction.
In the future, we will work on extending the model to solving dictionary elements that can be solved up to transformation which allows scaling the dictionary elements during the learning phase to handle assets that could be of varied different sizes and aspect ratios. This should make the model more robust to variation in the scale and size of the convolutional filters.

\bibliographystyle{splncs}
\bibliography{./source/egbib}
\end{document}